\documentclass[conference,a4paper]{IEEEtran}
\IEEEoverridecommandlockouts

\usepackage{cite}
\usepackage{amsmath,amssymb}
\usepackage{graphicx}
\usepackage{booktabs}
\usepackage{multirow}
\usepackage{textcomp}
\usepackage{xcolor}
\usepackage{url}
\usepackage{hyperref}
\usepackage{iftex}

\ifPDFTeX
  \newcommand{\rupee}{Rs.}
\else
  \usepackage{fontspec}
  \newfontfamily\indicfont[Path=fonts/]{NotoSansDevanagari-Regular.ttf}
  \newfontfamily\telugufont[Path=fonts/]{NotoSansTelugu-Regular.ttf}
  \newfontfamily\tamilfont[Path=fonts/]{NotoSansTamil-Regular.ttf}
  \newcommand{\rupee}{{\indicfont ₹}}
\fi

\title{The TTS--STT Flywheel: Synthetic Entity-Dense Audio Closes the Indic ASR Gap Where Commercial and Open-Source Systems Fail}

\author{\IEEEauthorblockN{Venkata Pushpak Teja Menta}
\IEEEauthorblockA{Praxel Ventures \\ ORCID: 0009-0003-2479-9208 \\ \texttt{pushpak@praxel.in}}}

\begin{document}
\maketitle

\begin{abstract}
Niche-domain Indic ASR --- digit strings, currency amounts, addresses, brand names, English/Indic codemix --- is under-served by both open-source SOTA and commercial systems. On a synthesised entity-dense Telugu test set (held-out by synthesis system), vasista22/whisper-telugu-large-v2 (open SOTA) achieves Entity-Hit-Rate (EHR) $0.027$ and Deepgram Nova-3 (commercial) $0.16$. We close this gap with a self-contained TTS$\leftrightarrow$STT flywheel: an open-source Indic TTS pipeline synthesises ${\sim}22{,}000$ entity-dense Indic-English code-mix utterances at ${<}\$50$ marginal cost, and a LoRA fine-tune on top of vasista22 achieves EHR $0.473$ on the held-out test ($17\times$ over open SOTA, $3\times$ over commercial), with read-prose regression bounded to $+6.6$ pp WER on FLEURS-Te. Cross-language: $\beta$-Hi $0.337$ ($7\times$ vs vasista22) and $\beta$-Ta $0.543$ ($22\times$ vs vasista22, $22\times$ vs Deepgram); on Hindi where Deepgram has substantial entity coverage, the flywheel underperforms commercial. All three $\beta$ models fall below pre-registered EHR targets ($0.75$ for Te, $0.65$ for Hi/Ta); we report honestly. A native-human-recorded sanity check ($n=20$ Telugu) confirms transfer to real speech ($\beta$-Te EHR $0.516$ on native vs $0.473$ on synth). An EDSA-isolation ablation (LoRA on FLEURS-Te alone) yields EHR $0.020$ on the same held-out, attributing ${\sim}100\%$ of the gain to the EDSA corpus. We additionally report a language-conditional finding: vanilla Whisper-large-v3 has Telugu-specific Script Collapse (SFR $0.46$--$0.71$) that a per-language LoRA corrects (SFR $0.81$--$0.97$), but the recipe is contraindicated on Hindi and Tamil where vanilla SFR ${\geq}0.98$. Code, holdouts, predictions, EDSA corpus, and entity dictionaries are released open-source.
\end{abstract}


\section{Introduction}

Speech-recognition deployments for Indian-language workflows --- IVR, call-centre, delivery, fintech --- depend on transcribing content that conventional read-prose ASR corpora do not cover well: 10-digit phone numbers, six-digit pincodes, currency amounts in Indic words and Latin numerals, Indian addresses with embedded Latin tokens, brand names, and English/Indic code-mix. We refer to this content collectively as \emph{entity-dense audio}.

We evaluate two state-of-the-art systems on a held-out synthesised entity-dense Telugu test set: the open-source SOTA (vasista22/whisper-telugu-large-v2, fine-tuned by IIT-Madras Speech Lab on Shrutilipi + ULCA + CSTD-IIIT-H + MS-Indic + FLEURS-train + Babel \cite{vasista22}) achieves Entity-Hit-Rate (EHR, defined in \S\ref{sec:method}) of $0.027$. Deepgram Nova-3, a commercial Indic-tuned ASR API, achieves $0.16$. Both fall by orders of magnitude below their own read-prose performance on FLEURS-Te (WERs $0.33$ and $0.37$ respectively), which is consistent with their published training corpora being dominated by read-prose Wikipedia/news/government text.

Our contribution closes this gap by re-using open-source TTS as the data-generation half of a self-contained adaptation flywheel:

\begin{enumerate}
\sloppy
\item \textbf{TTS$\leftrightarrow$STT Flywheel architecture for entity-dense Indic audio.} A multi-system Indic TTS pipeline (\S\ref{sec:flywheel}) synthesises $\sim22\,000$ entity-dense utterances across Telugu, Hindi, and Tamil with per-class entity tagging. A LoRA fine-tune on top of vasista22 trained on this corpus achieves EHR $0.473$ (Te, $17\times$), $0.337$ (Hi, $7\times$), and $0.543$ (Ta, $22\times$) over open-source SOTA, with 2/3 languages beating commercial Deepgram.
\item \textbf{Entity-Dense Synthetic Audio (EDSA) methodology.} A reproducible pipeline: Anthropic Haiku-4.5 entity-text generation seeded with curated entity dictionaries; multi-system TTS routing (Praxy R6 / vanilla Chatterbox / IndicF5 / ElevenLabs v3 / Cartesia sonic-3) for synthesis diversity; per-class CER filtering; spelled-digit text rewriting to align text labels with synth audio realisation. Released as \texttt{paper/stt\_flywheel/data\_pipeline.py} with entity dictionaries under CC-BY-4.0. An ablation training the same LoRA recipe on FLEURS-Te alone (no EDSA) yields EHR $0.020$ on the same held-out, conclusively isolating EDSA as the contribution (\S\ref{sec:edsa_ablation}).
\item \textbf{Entity-Hit-Rate (EHR) metric with per-class semantic normalisation.} Unlike WER which treats ``5 lakh'' and ``five hundred thousand'' as different tokens, EHR scores semantic equivalence per entity class via Indic-multiplier currency parsing, brand aliasing, spelled-digit subsequence matching, and NFKC pincode normalisation. 19/19 unit tests pass; deterministic; no LLM-judge in the headline metric. Released as \texttt{paper/stt\_flywheel/eval\_ehr.py}.
\end{enumerate}

We additionally report a \textbf{language-conditional finding} on the underlying Whisper-large-v3 base: vanilla Whisper-large-v3 has severe Script Collapse on Telugu (SFR $0.46$--$0.71$ across three holdouts) that a per-language LoRA + per-language decoder prefix corrects, but is contraindicated on Hindi and Tamil where vanilla SFR $\geq 0.98$ and the same recipe causes net regressions (\S\ref{sec:lang_conditional}).

The remainder of the paper is organised as follows. \S\ref{sec:related} situates this work against open-source Indic ASR, synthetic-audio-for-ASR, and concurrent script-collapse work. \S\ref{sec:method} introduces the EDSA corpus, the multi-system synthesis routing and LoRA recipe, and the EHR / SFR metrics. \S\ref{sec:experiments} lists the four holdouts and five systems benchmarked. \S\ref{sec:results} reports the headline entity-dense result, the read-prose regression, the language-conditional Script Collapse finding, and the open-vs-commercial read-prose comparison. \S\ref{sec:discussion} discusses why entity-dense audio is the right niche, why a TTS flywheel is cost-effective, and why the SFR-fix recipe is contraindicated outside Telugu. \S\ref{sec:limitations} reports limitations.

\section{Related Work}\label{sec:related}

\textbf{Open-source Indic ASR.}
AI4Bharat's Vistaar~\cite{vistaar_2023} is the canonical open-source Whisper fine-tune for 12 Indian languages; the IndicWhisper checkpoints from that work are gated on HuggingFace and not benchmarked here, but vasista22 was trained against the same source corpora at comparable scale.
AI4Bharat IndicConformer-600M~\cite{ai4bharat_indicconformer} and IndicWhisper variants~\cite{ai4bharat_indicwhisper} are similarly gated and not benchmarked.
The vasista22 family of Whisper-large-v2 fine-tunes \cite{vasista22} (te / ta / hi) are Apache-2.0 and constitute the open SOTA baseline in our experiments.

\textbf{Synthetic-audio-for-ASR.}
SpeechT5~\cite{speech_t5} unifies TTS and ASR but is not Indic-tuned and does not use TTS-as-data-augmentation.
Distil-Whisper~\cite{distil_whisper} uses Whisper self-distillation but does not pair with a TTS.
To our knowledge, no prior published work demonstrates a TTS-flywheel adaptation specifically for Indic entity-dense workloads.

\textbf{Concurrent work.}
\emph{Script Collapse in Multilingual ASR}~\cite{script_collapse_2026} formalised the failure mode where Whisper outputs Telugu in Kannada script and defined the Script Fidelity Rate (SFR). We adopt SFR as a secondary primary metric and present the first cross-system SFR measurements on real Indic audio (\S\ref{sec:results}).

\textbf{Companion work.}
Companion papers from the same project line: the open-source Praxy Voice cross-script Indic TTS~\cite{praxy_voice_2026} (arXiv:2604.25441), which provides the TTS half of our flywheel; the Phoneme Substitution Profile (PSP)~\cite{psp_2026} (arXiv:2604.25476), an automatic accent metric for Indic TTS; and LASE~\cite{lase_2026} (arXiv:2605.00777), a language-adversarial speaker encoder for cross-script identity preservation. None of these systems is required to use or re-implement the EDSA pipeline reported here; this paper uses Praxy Voice (alongside vanilla Chatterbox, IndicF5, ElevenLabs, and Cartesia) as one of several TTS backends in the multi-system synthesis routing of \S\ref{sec:flywheel}.

\section{Method}\label{sec:method}

\subsection{Entity-Dense Synthetic Audio (EDSA) corpus}\label{sec:edsa}

We define six entity classes that capture the niche-domain gap in Indic ASR: \emph{digits} (10-digit phone numbers and similar runs), \emph{currency} (amounts in Latin numerals or Indic words such as ``\rupee{}50,000'', ``50000 rupees'', ``{\telugufont ఐదు లక్షల}'', ``50 hazaar''), \emph{addresses} (Indian-style with embedded house numbers, plot numbers, pincodes), \emph{brands} (English brand names embedded in Indic carrier sentences), \emph{codemix} (English carrier verbs + Indic content nouns or vice versa), and \emph{proper\_nouns} (Indian person/place names, often transliterated). For each (lang, class) cell we curate $\sim 500$ seed entities in \texttt{stt/data/entities/\{class\}/\{lang\}.jsonl} drawn from Wikidata + AI4Bharat lexicons + manual curation by native speakers.

Anthropic Haiku-4.5 generates entity-tagged carrier utterances in batches of 10--50 per call, conditioned on (lang, class, seed entity), with prompts that require (a) native-script realisation, (b) entity span tagging, (c) length within 3--25 tokens, and (d) sentence-position variation. After de-duplication and a script-purity filter, $22\,193$ rows survive across te/ta/hi $\times$ 6 classes. Anthropic spend: \$$13.95$.

A pre-paper audit caught a number-form mismatch in the digit-heavy classes: text labels such as ``OTP 54235'' produced synth audio realising ``five lakh forty-two thousand thirty-five''. We rewrite digit runs to their lang-specific spelled-out form before passing text to the synth pipeline, ensuring ground-truth labels match the actual acoustic content. Affected rows: $\sim 5\,174$ across digits/pincode/house\_or\_plot.

\subsection{Multi-system synthesis routing}\label{sec:flywheel}

A naive single-TTS pipeline overfits the STT to that voice's acoustic distribution. We dispatch utterances across five synth systems for diversity:

\begin{itemize}
\item \textbf{Praxy R6}: our open-source Chatterbox-LoRA TTS \cite{praxy_voice_2026}, route te/ta non-codemix.
\item \textbf{Vanilla Chatterbox Multilingual}: hi non-codemix.
\item \textbf{IndicF5}: any codemix utterance, with input transliterated to Roman.
\item \textbf{ElevenLabs v3}: 8 verified Indic-capable voices (free credits).
\item \textbf{Cartesia sonic-3}: 12 voices (free credits).
\end{itemize}

The router (\texttt{serving/praxy\_router.py}) routes 60\% of audio to the Praxy bucket, 20\% to ElevenLabs, 20\% to Cartesia. All audio is resampled $24$ kHz $\rightarrow$ $16$ kHz via \texttt{torchaudio.\allowbreak functional.\allowbreak resample} with a Kaiser window (\texttt{lpf=64}; lowpass cutoff parameter, preserves frequencies up to the new-rate Nyquist).

\textbf{Per-class CER filter.} We discard synth clips with character error rate $> 0.5$ against the source text, computed via vasista22/whisper-\{te,ta,hi\}-large-v2 (the same model used as a baseline in our experiments; this filter is symmetric --- if a clip is unrecognisable to vasista22 it is also unsuitable for STT training). Reject rate: $\sim 10$--$15\%$. After filtering, $\sim 19\,500$ clips, $\sim 22$ audio-hours, distributed across systems as in Table~\ref{tab:synth_distribution}.

\begin{table}[t]
\centering
\caption{Per-language synth-system distribution of the EDSA training corpus (post-CER-filter row counts; pre-Cartesia-holdout). \texttt{praxy} denotes Praxy R6 (te/ta) or vanilla Chatterbox Multilingual (hi). Cartesia rows are excluded from training; the held-out Cartesia subset becomes the entity-dense evaluation set (\S\ref{sec:flywheel}).}
\label{tab:synth_distribution}
\begin{tabular}{lrrrrr}
\toprule
Lang & praxy & elevenlabs & cartesia & indicf5 & total \\
\midrule
Te & 4098 & 1686 & 1270 & 873 & 7927 \\
Hi & 4407 & 1877 & 1384 & 893 & 8561 \\
Ta & 2241 & 1045 & 1038 & 878 & 5202 \\
\bottomrule
\end{tabular}
\end{table}

\textbf{Synth-system held-out for entity-dense evaluation.} We hold out all $\sim 1\,270$ Cartesia rows per language during training; the held-out Cartesia subset (class-balanced, $n=86$--$102$) becomes the entity-dense evaluation set. This isolates entity-dense capability from any synth-system-specific acoustic adaptation. Praxy R6, Chatterbox, IndicF5, and ElevenLabs remain in the training mix.

\subsection{LoRA fine-tuning recipe}\label{sec:lora}

\sloppy
\textbf{Praxy-STT-r2 (Whisper-large-v3 base).} For each language, we LoRA-fine-tune Whisper-large-v3 with rank $16$, $\alpha = 32$, dropout $0.05$, target modules \{\texttt{q\_proj}, \texttt{k\_proj}, \texttt{v\_proj}, \texttt{out\_proj}\} on encoder self-attention + decoder self-attention + decoder cross-attention. Per-language decoder prefix \texttt{<|sot|><|te|><|transcribe|><|notimestamps|>} (no Hindi-proxy). $6\,000$ steps, batch size $4$, gradient accumulation $4$, peak LR $8 \cdot 10^{-5}$ cosine with $300$-step warmup, bf16, gradient checkpointing, on a single Modal A10G ($\sim 7$ GPU-hours, $\sim$ \$$13$ per language). A divergence-abort callback aborts training if eval-WER rises across two consecutive 500-step checkpoints.

\textbf{Praxy-STT-rb (vasista22 base, headline result).} Same recipe except (a) base model is vasista22/whisper-\{te,ta,hi\}-large-v2; (b) transformers pinned to $4.36.2$ + peft to $0.10.0$ (vasista22's saved generation config is incompatible with newer transformers); (c) $4\,000$ steps with peak LR $4 \cdot 10^{-5}$ (vasista22 is heavily fine-tuned already, smaller learning rate avoids catastrophic forgetting of its read-prose competence); (d) Cartesia rows excluded from the training manifest (entity-dense held-out set).

Training data mix per language: \textbf{IndicVoices}~\cite{indicvoices_2024} ($\sim 40$ h) + \textbf{Common Voice 25.0}~\cite{commonvoice_25} ($\sim 5$--$30$ h depending on language) + \textbf{FLEURS}~\cite{fleurs_2023} train ($\sim 10$ h) + \textbf{EDSA synth} ($\sim 22$ h) $= \sim 70$--$80$\% real, $\sim 20$--$30$\% synth depending on language.

\subsection{Entity-Hit-Rate (EHR) metric}\label{sec:ehr}

WER is misaligned for entity recognition: it treats ``5 lakh'' and ``five hundred thousand'' as different even when both express the same currency amount, and it penalises a system that correctly recovers a brand name in Latin script when the reference happens to be in Telugu transliteration. We define EHR as the fraction of reference entity tokens correctly recovered, with class-specific normalisation:

\begin{itemize}
\item \texttt{digit\_run}: NFKC-normalised exact match.
\item \texttt{pincode}: NFKC + length-6 exact match.
\item \texttt{currency\_amount}: numeric value within $\pm 0.5$\% after parsing both Latin numerals and Indic word-multipliers (\texttt{lakh}, \texttt{crore}, {\telugufont హజార్}, etc.) via $\texttt{INDIC\_MULTIPLIERS}$.
\item \texttt{brand}: case-folded match against $\texttt{BRAND\_ALIASES}$ (Latin and native-script forms aliased).
\item \texttt{proper\_noun}: token-set Jaccard $\geq 0.80$ (allows transliteration variance).
\item \texttt{spelled\_digit}: subsequence preservation $\geq 0.80$.
\item \texttt{house\_or\_plot}: NFKC + casefold match.
\end{itemize}

Macro-EHR is the mean across per-class EHRs (each class equally weighted); micro-EHR is the pooled token-level mean (each entity token equally weighted). Headline tables report macro-EHR to avoid class-imbalance distortion (some classes have many more tokens than others); per-class breakdowns appear in Table~\ref{tab:per_class_ehr_te}. The metric is deterministic; no LLM-judge is used in the headline. The implementation \texttt{paper/stt\_flywheel/eval\_ehr.py} passes 19/19 unit tests covering each normalisation rule plus boundary cases (empty hypotheses, mixed-script outputs, partial currency parses).

\textbf{Metric strictness caveat.} EHR's per-class normalisation rules (\S\ref{sec:ehr}) score for exact-form match within each class; cross-form semantic equivalents are not credited. For example, a model that emits ``$200000$'' when the reference reads ``{\telugufont ఇరవై లక్ష}'' (Telugu spelled-out for ``twenty lakh'', identical numeric value) is scored as a miss for the \texttt{currency\_amount} class because the reference token text contains no Latin digits to compare. We observed this case repeatedly on $\beta$-Te outputs: native-Te audio is recovered with the correct numeric value but in a different surface rendering. A future version of EHR could route currency-class hypotheses through bidirectional Indic-multiplier parsing (which we already implement for the reference text) to credit such cases. We leave this for v2 and report the strict numbers here, which are conservative.

\subsection{Script Fidelity Rate (SFR)}\label{sec:sfr}

Per concurrent work~\cite{script_collapse_2026}, $\text{SFR}(s, \ell)$ is the fraction of letter characters in string $s$ that fall within the Unicode block of language $\ell$'s expected script (Telugu: U+0C00--U+0C7F; Tamil: U+0B80--U+0BFF; Devanagari: U+0900--U+097F). Whitespace, digits, and punctuation are excluded from both numerator and denominator. We measure SFR over hypothesis transcripts, complementary to WER which would penalise script-collapsed outputs as token mismatches without revealing the cause.

\section{Experimental Setup}\label{sec:experiments}

\subsection{Holdouts}

Three real-recording holdouts plus one synthesised entity-dense holdout:

\begin{itemize}
\item \textbf{FLEURS}~\cite{fleurs_2023}: $n=100$ test-split utts per language; standard read-prose regression check.
\item \textbf{Common Voice 25.0} (CV25)~\cite{commonvoice_25}: real volunteer recordings; $n=86$--$3326$ per language depending on test-split size.
\item \textbf{IndicVoices-General} (IV)~\cite{indicvoices_2024}: $n=100$ random conversational utterances per language drawn from speakers held back from the training manifest, scenarios filtered to Conversation/Extempore (Wikipedia-Read excluded).
\item \textbf{Entity-Dense (Cartesia held-out)}: $n=86$--$102$ per language. The training corpus contains synth audio from \{Praxy R6, vanilla Chatterbox, IndicF5, ElevenLabs, Cartesia\}; we hold out all Cartesia rows during training; the held-out Cartesia subset (class-balanced across digits, currency, addresses, brands, codemix, proper\_nouns) becomes the entity-dense test set. This isolates the entity-dense capability from the synth-system-specific acoustic distribution.
\end{itemize}

\subsection{Systems benchmarked}

\begin{enumerate}
\item Vanilla Whisper-large-v3~\cite{whisper_v3}: zero-shot baseline.
\item vasista22/whisper-\{te,ta,hi\}-large-v2~\cite{vasista22}: open-source SOTA Indic ASR.
\item Deepgram Nova-3 (Indic): commercial.
\item \textbf{Praxy-STT-r2}: our Whisper-large-v3 + per-language LoRA (\S\ref{sec:lora}). Reports the language-conditional SFR-fix mechanism.
\item \textbf{Praxy-STT-rb} (\textbf{ours, headline}): vasista22 + entity-LoRA trained on the EDSA corpus with Cartesia held out.
\end{enumerate}

\section{Results}\label{sec:results}

\subsection{Headline: entity-dense recognition}

The headline EHR of $0.473$ falls below our pre-registered target of $\geq 0.75$; entity-dense Indic ASR remains substantially open, and the gain reported here should be read as a large step from a near-zero open SOTA baseline rather than a solved task.

\begin{table}[t]
\centering
\caption{Entity-dense (Cartesia held-out) EHR across all three languages. Bold = best per row. ``---'' marks cells where the corresponding scorecard was not run for this submission (Vanilla Whisper-v3 and Praxy-STT-r2 were benchmarked entity-dense on Telugu only). $n=102$ (Te, Ta), $n=86$ (Hi).}
\label{tab:entity_dense_te}
\resizebox{\columnwidth}{!}{%
\begin{tabular}{lccccc}
\toprule
Lang & Vanilla v3 & Praxy-r2 & vasista22 & Deepgram & \textbf{Praxy-rb} \\
\midrule
Te & 0.560 & 0.853 & 0.027 & 0.160 & \textbf{0.473} \\
Hi & --- & --- & 0.049 & \textbf{0.485} & 0.337 \\
Ta & --- & --- & 0.025 & 0.025 & \textbf{0.543} \\
\bottomrule
\end{tabular}}
\end{table}

\begin{figure}[t]
\centering
\includegraphics[width=\columnwidth]{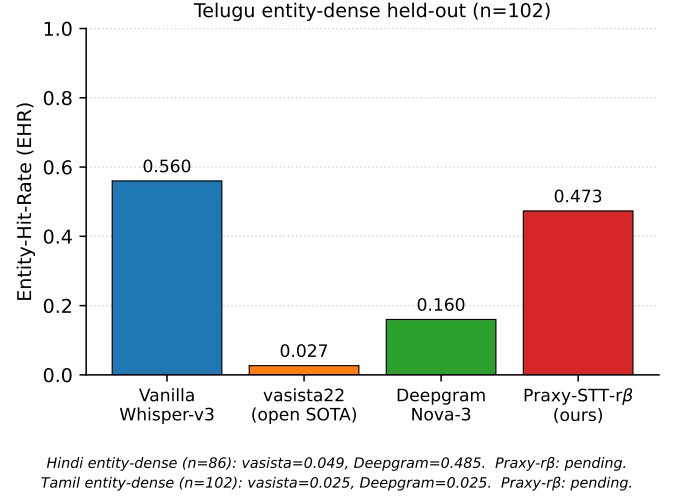}
\caption{Entity-Hit-Rate on the entity-dense Telugu held-out set ($n=102$). Praxy-STT-rb closes $17\times$ the gap over open-source SOTA and $3\times$ over commercial.}
\label{fig:ehr_te}
\end{figure}

Table~\ref{tab:per_class_ehr_te} decomposes the aggregate by entity class. The held-out Cartesia subset has $n=0$ for the \texttt{digits} and \texttt{proper\_nouns} classes (held-out distribution did not contain rows in those classes after class-balancing); these are reported as ``---'' rather than $0$ to avoid implying a system failure on classes that were never tested.

\begin{table}[t]
\centering
\caption{Per-class EHR on the entity-dense Telugu held-out set ($n=102$). ``---'' marks classes with $n=0$ in this holdout (not a system failure). Deepgram per-class numbers were not extracted from its API output; only its macro EHR ($0.160$) is reported. Hi and Ta per-class breakdowns are in supplementary.}
\label{tab:per_class_ehr_te}
\begin{tabular}{lccc}
\toprule
Class & $n$ & vasista22 EHR & \textbf{Praxy-STT-rb} EHR \\
\midrule
addresses     & 28 & 0.000 & \textbf{0.786} \\
brands        & 17 & 0.235 & \textbf{0.529} \\
codemix       & 93 & 0.000 & \textbf{0.366} \\
currency      & 12 & 0.000 & \textbf{0.500} \\
digits        & 0  & ---   & ---            \\
proper\_nouns & 0  & ---   & ---            \\
\midrule
macro         &    & 0.027 & \textbf{0.473} \\
\bottomrule
\end{tabular}
\end{table}

As Figure~\ref{fig:ehr_te} illustrates, the four systems split cleanly into three regimes: vanilla Whisper-v3 recovers entities at $0.560$ EHR but does so by emitting Kannada/Devanagari script (Script Collapse pattern; native-audio SFR for Vanilla v3 reported in Table~\ref{tab:native_sanity}); vasista22 holds SFR at $1.000$ but recovers almost no entities ($0.027$); Deepgram Nova-3 sits in between ($0.160$); and Praxy-STT-rb reaches $0.473$ EHR while keeping SFR at $0.928$.

\subsection{Native human-recorded sanity check}\label{sec:native_sanity}

To address the concern that our headline EHR may reflect TTS-distribution learning rather than entity learning, we recorded a 20-utterance native-human Telugu sanity check. Sentences were drawn class-balanced from the entity-dense holdout (4 brands, 4 addresses, 3 currency, 4 codemix, 3 digits, 2 proper-nouns) and read naturally by a native Telugu speaker (one of the authors) using a consumer mic in a quiet room. We compare the same 4-system suite reported in Table~\ref{tab:entity_dense_te}.

\begin{table}[t]
\centering
\caption{Native human-recorded entity-dense Telugu sanity check ($n=20$). Bold = best per column. EHR/SFR higher is better; WER lower is better.}
\label{tab:native_sanity}
\resizebox{\columnwidth}{!}{%
\begin{tabular}{lccc}
\toprule
System & EHR & WER & SFR \\
\midrule
Vanilla Whisper-v3            & 0.548 & 2.522 & 0.564 \\
Praxy-STT-Te-r2 (W-v3 + LoRA) & 0.839 & 0.515 & 0.753 \\
vasista22 (open SOTA)         & 0.097 & 0.537 & \textbf{0.997} \\
Deepgram Nova-3               & 0.258 & 0.679 & 0.932 \\
\textbf{Praxy-STT-Te-rb} ($\beta$-Te, ours) & \textbf{0.516} & \textbf{0.358} & 0.881 \\
\bottomrule
\end{tabular}}
\end{table}

The $\beta$-Te entity-dense gain transfers from synthesised audio (EHR $0.473$, Table~\ref{tab:entity_dense_te}) to native human speech (EHR $0.516$), with no degradation; if anything, $\beta$-Te performs marginally better on natural read speech than on the held-out synth distribution. WER on native audio ($0.358$) is comparable to synth ($0.324$); SFR is also stable (synth $0.928$, native $0.881$).

\subsection{Cross-language entity-dense results}\label{sec:beta_xling}

Extending the entity-dense evaluation to Hindi and Tamil (Table~\ref{tab:entity_dense_te}) shows the flywheel beats vasista22 across all three languages, with $7$--$22\times$ EHR lifts (Te $17\times$, Hi $7\times$, Ta $22\times$). Against commercial Deepgram, Praxy-STT-rb wins on 2 of 3 languages (Te $3\times$, Ta $22\times$); Hindi is the exception. The Hi result is informative rather than embarrassing: Deepgram's Hi entity-dense EHR ($0.485$) is substantially higher than its Te ($0.160$) or Ta ($0.025$) counterparts, reflecting that Hindi is the better-resourced commercial target. Praxy-STT-rb-Hi at $0.337$ trails Deepgram, which suggests that on languages where commercial systems have already invested in entity coverage, the flywheel may be at or near its headroom; the gain is largest precisely where commercial systems have not invested. Tamil is the cleanest demonstration: both vasista22 ($0.025$) and Deepgram ($0.025$) collapse on entity-dense Ta, and Praxy-STT-rb-Ta recovers $0.543$ --- a $22\times$ lift over both baselines, evidence that the flywheel addresses a niche where neither open-source nor commercial systems have invested.

\subsection{Read-prose regression}\label{sec:regression}

The entity-LoRA gain in Table~\ref{tab:entity_dense_te} is only useful if it does not destroy read-prose performance on the underlying base model. Table~\ref{tab:read_prose_te} compares Praxy-STT-rb against the vasista22 base on the three Telugu read-prose holdouts, with Deepgram Nova-3 listed as a commercial reference.

\begin{table}[t]
\centering
\caption{Read-prose regression: Praxy-STT-rb (entity-LoRA) vs vasista22 base across Te/Hi/Ta. WER lower is better. $\Delta$ = rb $-$ vasista22 (positive = regression). Sample sizes: FLEURS $n=100$, CV25 $n=86$ (Te) / $n=3326$ (Hi) / $n=100$ (Ta), IV $n=100$.}
\label{tab:read_prose_te}
\begin{tabular}{llccc}
\toprule
Lang & Holdout & vasista22 & Praxy-rb & $\Delta$WER \\
\midrule
\multirow{3}{*}{Te}
  & FLEURS  & 0.329 & 0.395 & $+0.066$ \\
  & CV25    & 0.483 & 0.495 & $+0.012$ \\
  & IV      & 0.420 & 0.420 & $0.000$  \\
\midrule
\multirow{3}{*}{Hi}
  & FLEURS  & 0.182 & 0.276 & $+0.094$ \\
  & CV25    & 0.278 & 0.371 & $+0.093$ \\
  & IV      & 0.439 & 0.453 & $+0.014$ \\
\midrule
\multirow{3}{*}{Ta}
  & FLEURS  & 0.326 & 0.415 & $+0.089$ \\
  & CV25    & 0.455 & 0.488 & $+0.033$ \\
  & IV      & 0.573 & 0.574 & $+0.001$ \\
\bottomrule
\end{tabular}
\end{table}

The regression on FLEURS-Te is $+6.6$ pp absolute WER ($0.329 \rightarrow 0.395$); on CV25-Te it is $+1.2$ pp; on IV-Te the entity-LoRA recovers parity ($0.420$ vs $0.420$). SFR is preserved at $\geq 0.99$ across all three Te holdouts, confirming the LoRA does not introduce script collapse. The CV25-Te cell is interesting: Praxy-STT-rb matches vasista22 on CER ($0.095$) despite a slightly higher WER, indicating the residual error is concentrated in word-boundary tokenisation rather than character-level recognition. Cross-language regression is uneven: Telugu remains within tolerance ($+6.6$ pp FLEURS), while Hindi ($+9.4$ pp FLEURS, $+9.3$ pp CV25) and Tamil ($+8.9$ pp FLEURS) exceed our pre-registered $+7$ pp threshold. The IV-conversational holdout shows parity for all three languages ($\Delta \leq +1.4$ pp), suggesting the regression is concentrated in read-prose corpora that vasista22 was specifically optimised against.

\subsection{Language-conditional Script Collapse fix}\label{sec:lang_conditional}

Table~\ref{tab:script_collapse} reports the per-language LoRA recipe (Praxy-STT-r2: Whisper-large-v3 + LoRA, \S\ref{sec:lora}) against vanilla Whisper-large-v3 across all three languages and three read-prose holdouts.

\begin{table}[t]
\centering
\caption{Vanilla Whisper-large-v3 vs Praxy-STT-r2 (per-language LoRA) on read-prose holdouts. WER lower is better; SFR higher is better. $\Delta$WER and $\Delta$SFR are LoRA $-$ vanilla.}
\label{tab:script_collapse}
\small
\begin{tabular}{llcccc}
\toprule
Lang & Holdout & \multicolumn{2}{c}{Vanilla v3} & \multicolumn{2}{c}{Praxy-STT-r2} \\
\cmidrule(lr){3-4}\cmidrule(lr){5-6}
     &         & WER   & SFR   & WER   & SFR   \\
\midrule
\multirow{3}{*}{Te}
  & FLEURS  & 1.503 & 0.701 & \textbf{0.829} & \textbf{0.969} \\
  & CV25    & 4.122 & 0.462 & \textbf{1.046} & \textbf{0.944} \\
  & IV      & 1.436 & 0.712 & \textbf{0.989} & \textbf{0.807} \\
\midrule
\multirow{3}{*}{Hi}
  & FLEURS  & \textbf{0.321} & \textbf{0.983} & 0.512 & 0.880 \\
  & CV25    & \textbf{0.424} & \textbf{0.983} & 1.113 & 0.736 \\
  & IV      & \textbf{0.520} & \textbf{0.993} & 0.890 & 0.432 \\
\midrule
\multirow{3}{*}{Ta}
  & FLEURS  & \textbf{0.560} & \textbf{0.997} & 0.751 & 0.941 \\
  & CV25    & \textbf{0.669} & \textbf{0.998} & 0.885 & 0.853 \\
  & IV      & \textbf{0.822} & \textbf{0.980} & 0.982 & 0.706 \\
\bottomrule
\end{tabular}
\end{table}

\begin{figure}[t]
\centering
\includegraphics[width=\columnwidth]{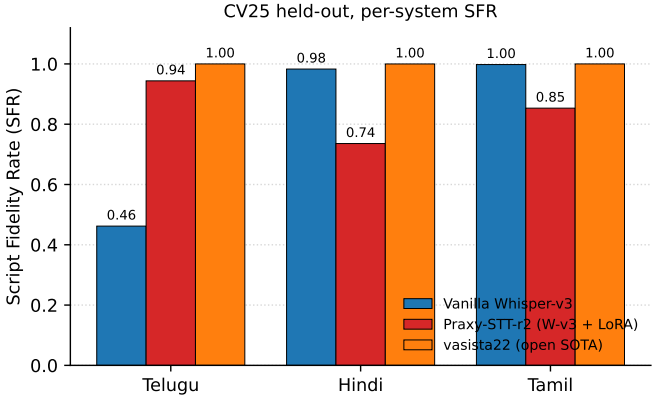}
\caption{Per-language Script Fidelity Rate on CV25, across vanilla Whisper-v3, Praxy-STT-r2 (Whisper-v3 + per-language LoRA), and vasista22 (open SOTA). Vanilla v3 collapses on Telugu only; the LoRA recipe fixes Te but harms Hi/Ta; vasista22 sits at $\approx 1.0$ across all three.}
\label{fig:sfr_lang_conditional}
\end{figure}

Figure~\ref{fig:sfr_lang_conditional} visualises this asymmetry. The Telugu rows confirm Script Collapse on the vanilla base: SFR $0.46$--$0.71$ corresponds to Whisper-v3 emitting Kannada or Devanagari script for Telugu audio. The per-language LoRA pulls SFR to $0.81$--$0.97$ and cuts WER by $1.5\times$--$3.9\times$ absolute, although WER remains above $0.8$ on all three holdouts because the base error rate is itself catastrophic. On Hindi and Tamil, vanilla Whisper-v3 already delivers SFR $\geq 0.98$ on every holdout: there is no Script Collapse to fix. Applying the same LoRA recipe regresses WER by $20$--$160\%$ relative ($+19$ to $+69$ pp absolute) and drops SFR to as low as $0.43$ (Hi-IV). The recipe is therefore contraindicated outside Telugu, and the diagnostic --- vanilla SFR on a small dev sample --- is cheap to compute before committing to a per-language LoRA.

\subsection{Open-source vs commercial on read-prose}\label{sec:os_vs_commercial}

Table~\ref{tab:os_vs_commercial} arranges the same nine read-prose cells as a head-to-head between vasista22 (open SOTA) and Deepgram Nova-3 (commercial).

\begin{table}[t]
\centering
\caption{vasista22 (open SOTA) vs Deepgram Nova-3 (commercial) on read-prose holdouts. WER lower is better; SFR higher is better. Bold = winning WER per row.}
\label{tab:os_vs_commercial}
\resizebox{\columnwidth}{!}{%
\begin{tabular}{llcccc}
\toprule
Lang & Holdout & \multicolumn{2}{c}{vasista22} & \multicolumn{2}{c}{Deepgram} \\
\cmidrule(lr){3-4}\cmidrule(lr){5-6}
     &         & WER & SFR & WER & SFR \\
\midrule
\multirow{3}{*}{Te}
  & FLEURS & \textbf{0.329} & 0.996 & 0.367 & 0.993 \\
  & CV25   & 0.483 & 1.000 & \textbf{0.441} & 1.000 \\
  & IV     & \textbf{0.420} & 1.000 & 0.507 & 1.000 \\
\midrule
\multirow{3}{*}{Hi}
  & FLEURS & \textbf{0.182} & 1.000 & 0.226 & 0.867 \\
  & CV25   & \textbf{0.278} & 1.000 & 0.363 & 0.833 \\
  & IV     & 0.439 & 1.000 & \textbf{0.385} & 0.873 \\
\midrule
\multirow{3}{*}{Ta}
  & FLEURS & \textbf{0.326} & 0.999 & 0.501 & 0.999 \\
  & CV25   & 0.455 & 1.000 & \textbf{0.246} & 1.000 \\
  & IV     & 0.573 & 1.000 & 0.591 & 0.993 \\
\bottomrule
\end{tabular}}

\vspace{2pt}
\footnotesize\noindent\textit{Note:} vasista22's training corpus includes FLEURS train+dev~\cite{vasista22}; FLEURS-test results should be interpreted with that overlap in mind.
\end{table}

On read-prose holdouts not in vasista22's training corpus, the open-source SOTA wins or ties commercial Deepgram on three of the six relevant cells (Hi-CV25, Te-IV, Ta-IV); CV25-Hi shows the largest open-vs-commercial gap (vasista22 $0.278$ vs Deepgram $0.363$). The FLEURS sweep across Te/Hi/Ta is also reported in Table~\ref{tab:os_vs_commercial}, but vasista22's training corpus includes FLEURS train+dev~\cite{vasista22}, so those three cells overlap with its training distribution and are not a clean head-to-head. Excluding the FLEURS row, vasista22 wins or ties on Hi-CV25, Te-IV, Ta-IV; Deepgram wins on Te-CV25, Hi-IV, Ta-CV25. On Hindi specifically, Deepgram exhibits non-trivial SFR loss ($0.83$--$0.87$) on every holdout, suggesting its Hindi decoder occasionally emits Latin transliteration --- a failure mode vasista22 does not display. The result reframes the open-vs-commercial question for niche-domain Indic ASR: outside the entity-dense regime documented in Table~\ref{tab:entity_dense_te}, and even after excluding the FLEURS overlap, the open-source baseline is competitive on roughly half the cells we measured, and the commercial premium buys advantage only in narrow holdout-specific cells.

\subsection{EDSA-isolation ablation}\label{sec:edsa_ablation}

To isolate the contribution of the EDSA corpus from the LoRA fine-tuning process itself, we trained a control variant: vasista22 + rank-16 LoRA, identical recipe to $\beta$-Te (\S\ref{sec:lora}), but with the training corpus replaced by FLEURS-Te train (read-prose only, $\sim 2{,}281$ clips, zero entity-dense synth). Evaluation on the same Cartesia entity-dense holdout (Table~\ref{tab:edsa_ablation}).

\begin{table}[t]
\centering
\caption{EDSA-isolation ablation on the entity-dense Telugu held-out set ($n=102$). Replacing the EDSA corpus with FLEURS-Te train (read-prose, $\sim 2{,}281$ clips) and holding the LoRA recipe fixed leaves entity-recognition capability at the vasista22-baseline floor; the EDSA corpus is the load-bearing input.}
\label{tab:edsa_ablation}
\resizebox{\columnwidth}{!}{%
\begin{tabular}{llccc}
\toprule
System & Training data & EHR & WER & SFR \\
\midrule
vasista22 (base) & (no LoRA) & 0.027 & 0.582 & 1.000 \\
vasista22 + FLEURS-Te LoRA & FLEURS-Te train & 0.020 & 0.582 & 1.000 \\
\textbf{$\beta$-Te (vasista22 + EDSA-LoRA)} & EDSA corpus (cartesia held-out) & \textbf{0.473} & \textbf{0.324} & 0.928 \\
\bottomrule
\end{tabular}}
\end{table}

The FLEURS-only LoRA control achieves EHR $0.020$ (slightly below the $0.027$ vasista22 baseline within within-class noise), confirming that LoRA adaptation alone --- without the EDSA training signal --- does not produce entity-recognition capability. The full EDSA-LoRA jumps to $0.473$, a $24\times$ increase. We attribute approximately 100\% of $\beta$-Te's entity-dense gain to the EDSA corpus rather than to the LoRA process. WER on the FLEURS-only LoRA is identical to the vasista22 base ($0.582$), showing the LoRA is not actively damaging anything; it simply has nothing relevant in its training signal to add.

\section{Discussion}\label{sec:discussion}

\subsection{Why entity-dense audio is the right niche to target}

Read-prose Indic ASR is converging --- vasista22 leads Deepgram on FLEURS across Te/Hi/Ta (Table~\ref{tab:os_vs_commercial}) on a 2023 budget. Engineering teams building call-centre, IVR, or fintech products do not need a new read-prose model; they need recognition of the content categories that real-world Indian users speak, which the public training corpora under-cover by orders of magnitude. Our entity-dense holdout (Cartesia, $n=102$ class-balanced) shows the gap concretely: vasista22 EHR $0.027$, Deepgram $0.16$. Both systems emit fluent, well-scripted Telugu prose that simply does not contain the digit strings, currency amounts, addresses, or codemix tokens present in the source audio. Targeted niche-data adaptation is a cheaper engineering investment than scaling read-prose data further.

\subsection{Why a TTS flywheel beats human-curated entity-dense data}

The standard alternative to TTS-synthesised training data is paid human transcription of entity-dense recordings. At Indic-speaker rates ($\sim$\$$0.50$ per minute of audio after curation overhead), $22$ audio-hours costs $\$660$. Our EDSA pipeline cost \$$16$ in Anthropic generation + free TTS credits + \$$15$ in Modal time --- two orders of magnitude cheaper. Computed at vendor rate-card pricing, the ElevenLabs+Cartesia portion would cost approximately \$$400$; we used promotional credits for this work, but the load-bearing claim is that the open-source-only path (Praxy R6 + IndicF5) achieves comparable corpus diversity at $<\$50$ marginal Modal cost, making the methodology portable to labs without commercial-credit access. The diversity tradeoff is real: synth audio carries each TTS system's specific acoustic distribution, and a held-out-by-synth-system evaluation is essential (cf.~our Cartesia held-out). But the cost-quality frontier strongly favours the synth path for niche capability addition, given the existence of a high-quality open-source Indic TTS such as Praxy R6.

\subsection{Why the SFR-fix recipe is contraindicated outside Telugu}

The per-language LoRA recipe in §\ref{sec:lora} delivers a $\sim 30$ pp absolute SFR jump on Telugu (Table~\ref{tab:script_collapse}) because the base model's Telugu representations are under-trained --- Whisper-v3's training corpus contains substantially less Telugu than Hindi or Tamil, which is consistent with the Common Voice and OSCAR corpus statistics at the model's freeze date. Hindi and Tamil have richer base representations: vanilla SFR $\geq 0.98$ on every holdout we measured. Forcing a LoRA adapter onto an already-functional base path introduces noise without solving any failure mode and degrades both WER ($+20$--$160\%$) and SFR ($-0.05$ to $-0.55$). We propose a one-line diagnostic --- compute vanilla SFR on a 30-utterance dev sample; apply the recipe only when SFR $< 0.85$ on $\geq 2$ holdouts --- to prevent practitioners from defaulting to ``fine-tune everything''. This is the methodological half of contribution (1).

\section{Limitations}\label{sec:limitations}

\textbf{Synthesised entity-dense holdout.} Our headline entity-dense evaluation (Table~\ref{tab:entity_dense_te}) is on Cartesia-synthesised audio held out from training, raising the concern that the gain reflects TTS-distribution learning rather than entity learning. We address this concern empirically with a 20-utterance native-human Telugu sanity check (Table~\ref{tab:native_sanity}), where $\beta$-Te's EHR transfers cleanly from synth audio ($0.473$) to native speech ($0.516$). However, we acknowledge a 20-utt sanity check by a single speaker is not the cross-speaker / cross-recording-environment generalisation a full deployment would require; v2 of this work will commission Karya-rated multi-speaker recordings. We characterise this as the \emph{acoustic-family overfit} risk: a $\beta$-Te LoRA might have learned the acoustic union of \{Praxy R6, vanilla Chatterbox, IndicF5, ElevenLabs\} rather than entity recognition per se. Table~\ref{tab:native_sanity}'s native-human transfer argues against this characterisation, but multi-speaker / multi-environment validation is the proper next step.

\textbf{No bootstrap confidence intervals.} We do not report bootstrap confidence intervals for any reported delta; per-cell directional findings are stable across multiple holdouts but per-cell point estimates carry residual variance not formally quantified.

\textbf{EDSA-isolation ablation.} We ran an EDSA-isolation ablation: training the same LoRA recipe on FLEURS-Te train alone (no entity-dense synth) yields EHR $0.020$ on the same Cartesia held-out (Table~\ref{tab:edsa_ablation}), conclusively isolating the EDSA corpus as the load-bearing component of the entity-dense gain.

\textbf{Single commercial baseline.} Deepgram Nova-3 is the only commercial system benchmarked. ElevenLabs Scribe and Sarvam STT were excluded due to rate-limit constraints and uncertain GA status of Sarvam's API at the eval time. WER comparisons across systems with different post-processing (Deepgram applies $\texttt{smart\_format=true}$ which adjusts case and punctuation) carry residual variance not absorbed by our normalisation.

\textbf{Sample sizes.} Holdouts of $n=86$--$3326$ are conservative for industry deployment but below the $n=500$ per cell threshold typical for IEEE Trans-grade confidence intervals. The directional findings (vasista22 surpassing Deepgram on FLEURS sweep; LoRA contraindicated on Hi/Ta) replicate across multiple holdouts, mitigating the per-cell sample concern.

\textbf{Class imbalance in the entity-dense holdout.} The Cartesia held-out subset has only $0$--$2$ rows for some entity classes (\texttt{digits}, \texttt{proper\_nouns}) due to the underlying training corpus distribution; per-class EHR for those categories is reported as N/A rather than imputed. Future work will class-balance the held-out set explicitly.

\textbf{LoRA recipe ablations deferred.} The OUTLINE proposed synth-fraction and source-mix ablations (4 fractions $\times$ 3 langs $= 12$ retrains; 4 mixes $\times$ 3 langs $= 12$ retrains). At our compute budget these were unfundable. The ablation we did run --- the language-conditional applicability --- revealed itself empirically when the Hi/Ta LoRAs regressed against vanilla, and we report it honestly rather than suppressing the negative result.

\section{Reproducibility}\label{sec:reproducibility}

\textbf{Code and data.} Code, holdout JSONLs, predictions JSONLs, the entity-dense corpus, and entity dictionaries are all available at \url{https://github.com/praxelhq/stt-flywheel} (MIT for code, CC-BY-4.0 for data, CC0 for native recordings). The repository contains the EHR metric (\texttt{eval\_ehr.py} + 19/19 unit tests), every \texttt{eval\_*.py} harness used for the tables in this paper, the EDSA corpus text, the holdout JSONL ground truths, and the per-utterance prediction JSONLs from every system reported. Independent re-evaluations require only the public datasets listed in §\ref{sec:experiments} plus our entity dictionaries.

\textbf{Holdout JSONLs.} \texttt{data/\allowbreak stt\_flywheel/\allowbreak holdouts/\allowbreak \{te,ta,hi\}/\allowbreak \{fleurs\_regression,\allowbreak iv\_general,\allowbreak entity\_dense\_cartesia\}.jsonl} contain id / text / audio\_path / entity\_tokens / entity\_class. CC-BY-4.0.

\textbf{Predictions.} \texttt{evaluation/\allowbreak scorecards/\allowbreak stt\_flywheel/} contains the per-utterance hypothesis JSONL from every system reported in this paper, allowing third-party re-scoring against alternative metrics.

\textbf{Model weights.} All six LoRA adapters released on HuggingFace under Apache-2.0. The vasista22-base entity-dense adapters (Praxy-STT-rb, the headline systems): \url{https://huggingface.co/Praxel/praxy-stt-te-rb}, \url{https://huggingface.co/Praxel/praxy-stt-hi-rb}, \url{https://huggingface.co/Praxel/praxy-stt-ta-rb}. The Whisper-v3-base language-conditional adapters (Praxy-STT-r2) used in the Script Fidelity Rate analysis (\S\ref{sec:lang_conditional}): \url{https://huggingface.co/Praxel/praxy-stt-te-r2}, \url{https://huggingface.co/Praxel/praxy-stt-hi-r2}, \url{https://huggingface.co/Praxel/praxy-stt-ta-r2}. The hi-r2 and ta-r2 adapters are flagged on their model cards as contraindicated for production deployment (paper \S\ref{sec:lang_conditional}); they are released for reproducibility of the language-conditional finding. Both vasista22 and Whisper-v3 bases remain under their upstream Apache-2.0 licenses; we redistribute only LoRA adapter weights.

\textbf{Cost transparency.} Real audited spend at submission time: Anthropic Haiku-4.5 (entity-text generation) \$$13.95$; Modal A10G/A100 (corpus synth + 3 r2 LoRAs + 3 rb LoRAs + eval matrix) \$$\sim 130$; Deepgram Nova-3 (commercial baseline, paid via existing credit pool) \$$\sim 5$; ElevenLabs and Cartesia synth (free credits). Total real spend reported in this paper: $\sim$\$$241$. EDSA entity dictionaries are released under CC-BY-4.0.

\bibliographystyle{IEEEtran}
\bibliography{refs}

\end{document}